\documentclass[letterpaper]{article} 
\usepackage{aaai2026}  
\usepackage{times}  
\usepackage{helvet}  
\usepackage{courier}  
\usepackage[hyphens]{url}  
\usepackage{graphicx} 
\usepackage{natbib}  
\usepackage{caption} 
\usepackage{algorithm}
\usepackage{algorithmic}

\usepackage{newfloat}
\usepackage{listings}
\DeclareCaptionStyle{ruled}{labelfont=normalfont,labelsep=colon,strut=off} 
\floatstyle{ruled}
\newfloat{listing}{tb}{lst}{}
\floatname{listing}{Listing}
\usepackage{times}
\usepackage{epsfig}
\usepackage{graphicx}
\usepackage{amsmath}
\usepackage{amssymb}
\usepackage{enumitem}
\usepackage{bm}
\usepackage{booktabs}
\usepackage{multirow}
\usepackage{makecell}
\usepackage{array}
\usepackage{colortbl}
\usepackage{bbding}
\usepackage{pifont}
\usepackage{subcaption}
\usepackage{blindtext}

\usepackage{algorithm}
\usepackage{algorithmic}
\usepackage{listings}

\usepackage{xspace}

\newcommand{\eg}{\textit{e.g.}\xspace}

\newcommand{\ie}{\textit{i.e.}\xspace}

\newcommand{\etal}{\textit{et al.}\xspace}

\definecolor{graycolor}{rgb}{0.95,0.95,0.95}

\definecolor{redcolor}{rgb}{0.67,0.23,0.21}
\definecolor{greencolor}{rgb}{0.28,0.61,0.36}

\def\textred#1{\textcolor{black}{#1}}
\def\textgreen#1{\textcolor{black}{#1}}

\def\textgreenbf#1{\textbf{\textcolor{black}{#1}}}

\newcounter{checksubsection}
\newcounter{checkitem}[checksubsection]

\title{Real Noise Decoupling for Hyperspectral Image Denoising}
\author{
    Yingkai Zhang\textsuperscript{\rm 1},
    Tao Zhang\textsuperscript{\rm 2},
    Jing Nie\textsuperscript{\rm 1},
    Ying Fu\textsuperscript{\rm 1}\thanks{Corresponding author}
}
\affiliations{
    \textsuperscript{\rm 1}Beijing Institute of Technology, Beijing, China\\
    \textsuperscript{\rm 2}Hangzhou Dianzi University, Hangzhou, China\\
    zhangyingkai@bit.edu.cn, tzhang@hdu.edu.cn, \{3420235028, fuying\}@bit.edu.cn

}

\usepackage{bibentry}

\begin{document}

\maketitle

\begin{abstract}
Hyperspectral image (HSI) denoising is a crucial step in enhancing the quality of HSIs. Noise modeling methods can fit noise distributions to generate synthetic HSIs to train denoising networks. However, the noise in captured HSIs is usually complex and difficult to model accurately, which significantly limits the effectiveness of these approaches.
In this paper, we propose a \textbf{multi-stage noise-decoupling framework} that decomposes complex noise into explicitly modeled and implicitly modeled components. This decoupling reduces the complexity of noise and enhances the learnability of HSI denoising methods when applied to real paired data.
Specifically, for \textbf{explicitly modeled noise}, we utilize an existing noise model to generate paired data for pre-training a denoising network, equipping it with prior knowledge to handle the explicitly modeled noise effectively.
For \textbf{implicitly modeled noise}, we introduce a high-frequency wavelet guided network. Leveraging the prior knowledge from the pre-trained module, this network adaptively extracts high-frequency features to target and remove the implicitly modeled noise from real paired HSIs.
Furthermore, to effectively eliminate all noise components and mitigate error accumulation across stages, a \textbf{multi-stage learning strategy}, comprising separate pre-training and joint fine-tuning, is employed to optimize the entire framework.
Extensive experiments on public and our captured datasets demonstrate that our proposed framework outperforms state-of-the-art methods, effectively handling complex real-world noise and significantly enhancing HSI quality.
\end{abstract}

\begin{links}
    \link{Code}{https://github.com/BITYKZhang/RND}
\end{links}

\section{Introduction}

Hyperspectral image (HSI) provides a wealth of spectral information, making it indispensable in a variety of applications, \eg, remote sensing~\cite{blackburn2007hyperspectral, thenkabail2016hyperspectral, zhang2010object}, classification~\cite{azar2020hyperspectral, cao2019hyperspectral}, and recognition~\cite{pan2003face, kim20123d}. Despite the potential of HSI, its quality is often compromised by various real-world noises due to the limitations of imaging technology and the complexity of the capture environment. 
Therefore, a robust and effective denoising solution is essential.

Recently, deep learning has emerged as a powerful alternative for HSI denoising, enabling the direct learning of mappings from noisy to clean images~\cite{zhang2023hyperspectral,lai2023hybrid, pang2024hir,xiao2024bridging,xiao2024region}. 
Despite the remarkable progress of learning-based methods, they still face learning bottlenecks. The complex mapping from real-world noisy images to their clean counterparts, driven by intricate noise distributions and the scarcity of high-quality paired training data, remains challenging to learn, thereby limiting their denoising performance on real-world HSIs.

To tackle these challenges, particularly the scarcity of data, one prominent line of research focuses on noise modeling to synthesize realistic HSIs~\cite{zhang2022guided}.
However, real-world noise is complex and difficult to model accurately, creating a domain gap between synthetic and real data that undermines the effectiveness of these methods. 
This difficulty persists even for well-understood types like readout and stripe noise, often due to inaccurate parameter fitting or the presence of unknown noise components. As illustrated in Figure~\ref{fig: Noise Motivation}, this discrepancy is significant: visually, many regions in the real noise cannot be properly fitted by the model (d-f), and quantitatively, there are notable PSNR differences between the real and synthetic HSIs. 
This domain gap severely degrades the quality of the synthetic data, making it challenging for models to learn mappings that generalize effectively to real-world scenarios.

To address these limitations, we propose a novel \textbf{multi-stage noise-decoupling framework}. Our key idea is to decompose complex real-world noise into two parts: an \textbf{explicitly modeled} component that can be described by physical noise models, and an \textbf{implicitly modeled} component, which includes residuals from inaccurate fitting and other unknown noise sources. By tackling these components separately, our framework simplifies the learning task at each stage and enhances the overall denoising performance.
Specifically, for \textbf{explicitly modeled noise}, we pre-train a state-of-the-art (SOTA) denoising network on synthetic data generated from an SOTA noise model. We aim to equip it with prior knowledge for handling these well-defined noise patterns, which is essential for the subsequent decoupling step.
Furthermore, for \textbf{implicitly modeled noise} (\ie, poorly fitted or unknown components), we introduce a high-frequency wavelet guided network motivated by the high correlation in the high-frequency of residuals between real and synthetic noise (Figures~\ref{fig: Noise Motivation}(a-c)). Leveraging the information-preserving nature of the wavelet transform, this network adaptively extracts high-frequency noise features from spectral-spatial information. By utilizing the prior knowledge from the pre-trained denoising network, the high-frequency wavelet guided network is capable of removing the implicitly modeled noise decoupled from the real complex noise, based on real paired data.
To effectively eliminate all noise components and mitigate error accumulation across stages, we develop a \textbf{multi-stage learning strategy} to guide the network. The strategy consists of three stages: 1) pre-training for explicitly modeled noise removal, 2) learning for implicitly modeled noise removal, and 3) fine-tuning for noise accumulation error and whole noise removal. The strategy focuses on removing noise, improving spectral fidelity, and alleviating noise accumulation error.
Through the synergy of our multi-stage network and learning strategy, our noise-decoupling framework robustly removes complex real-world noise and is not limited to improving upon the existing SOTA noise models and denoising networks. Extensive experiments on public and our captured datasets demonstrate the superior performance of our approach compared with other SOTA methods.
In summary, the contributions of our paper are as follows:
\begin{itemize}
   \item We propose a multi-stage noise-decoupling framework to effectively disentangle and remove complex noise components, including explicitly and implicitly modeled noise, based on the real paired data.
   \item We introduce a high-frequency wavelet guidance for adaptively extracting high-frequency features to suppress implicitly modeled noise.
   \item We develop a multi-stage learning strategy to remove the real-world noise by separating pre-training and joint fine-tuning, thereby mitigating the noise accumulation error.
\end{itemize}

\section{Related Work}

In this section, we review the two most relevant areas of work: HSI denoising methods and noise modeling methods.

\subsection{HSI Denoising Methods}

%
%

HSI denoising is a fundamental task in hyperspectral image processing, aiming to remove noise components while preserving the underlying clean information~\cite{he2019non, shi2021hyperspectral}. Existing methods can be broadly categorized into two groups: traditional optimization-based methods and deep learning-based methods.

Traditional optimization-based methods typically formulate the denoising task as an optimization problem, which is addressed by imposing various handcrafted regularizations, \eg, non-local similarity~\cite{fu2017adaptive, maggioni2012nonlocal}, low-rankness~\cite{wei2019low, sun2017hyperspectral,  zhao2020fast}, and spatial or spectral total variation priors~\cite{peng2020enhanced, yuan2012hyperspectral}. However, these optimization-based methods rely on handcrafted priors, which cannot sufficiently represent the nonlinearity and complexity of various realistic HSIs.

Recently, deep learning has developed rapidly~\cite{zhang2025unaligned, tian2023transformer, jiang2024towards} and has been applied to learning denoising mappings in a purely data-driven manner in several works~\cite{li2023spectral,xiao2024bridging,xiao2024region,zhang2024deep,zou2025calibration}.
Despite significant advancements in network design, existing learning-based methods still face challenges in handling complex data mappings due to intricate noise distributions and the scarcity of real data, which ultimately degrade their denoising performance.

\begin{figure}[t]
   \begin{center}
   \includegraphics[width=1\linewidth]{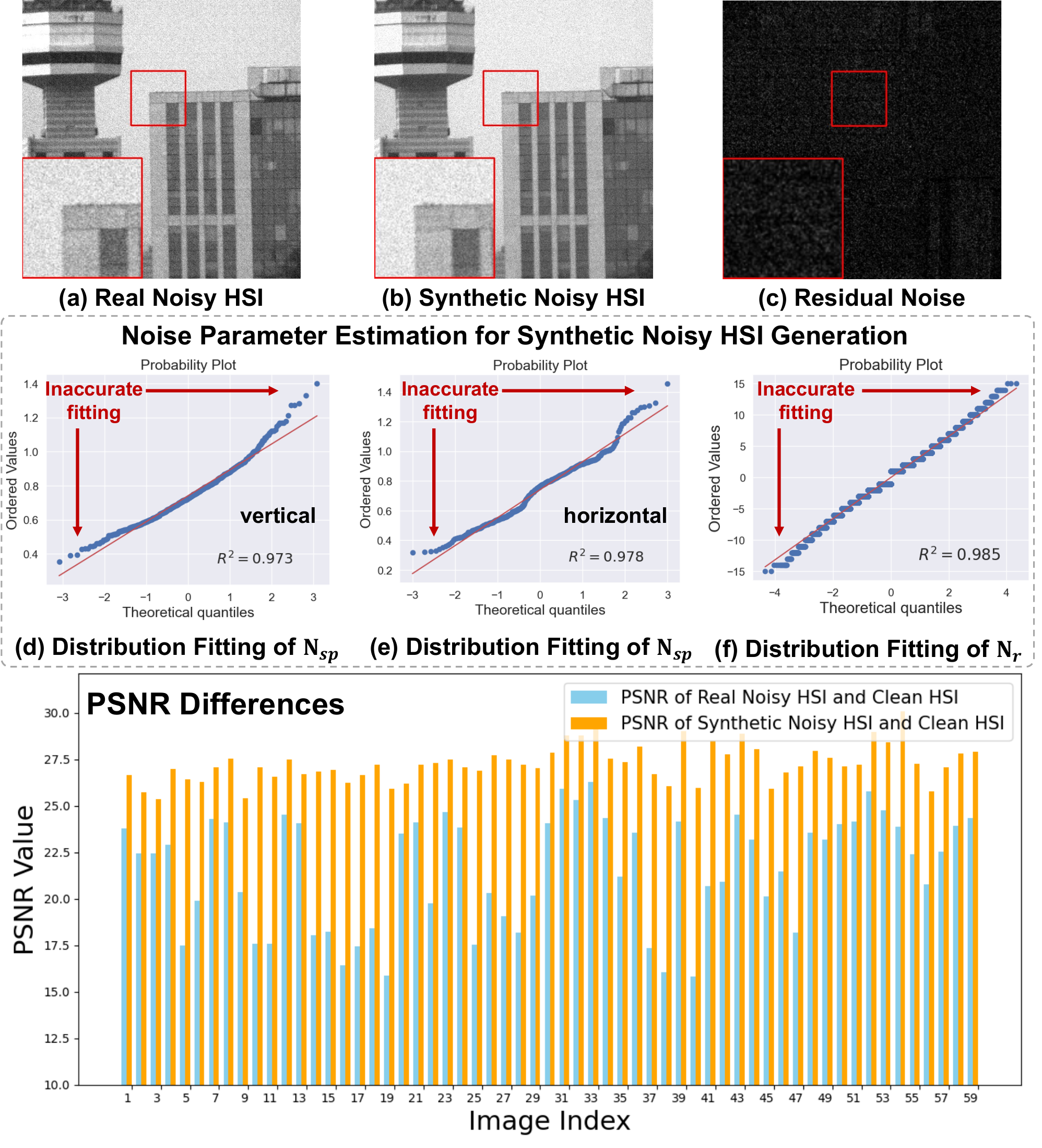}
   \end{center}
      \caption{The significant differences between real noisy HSIs and synthetic noisy HSIs generated by the noise model~\cite{zhang2022guided}.}
   \label{fig: Noise Motivation}
   \end{figure}

\begin{figure*}[t]
   \begin{center}
   \includegraphics[width=1\linewidth]{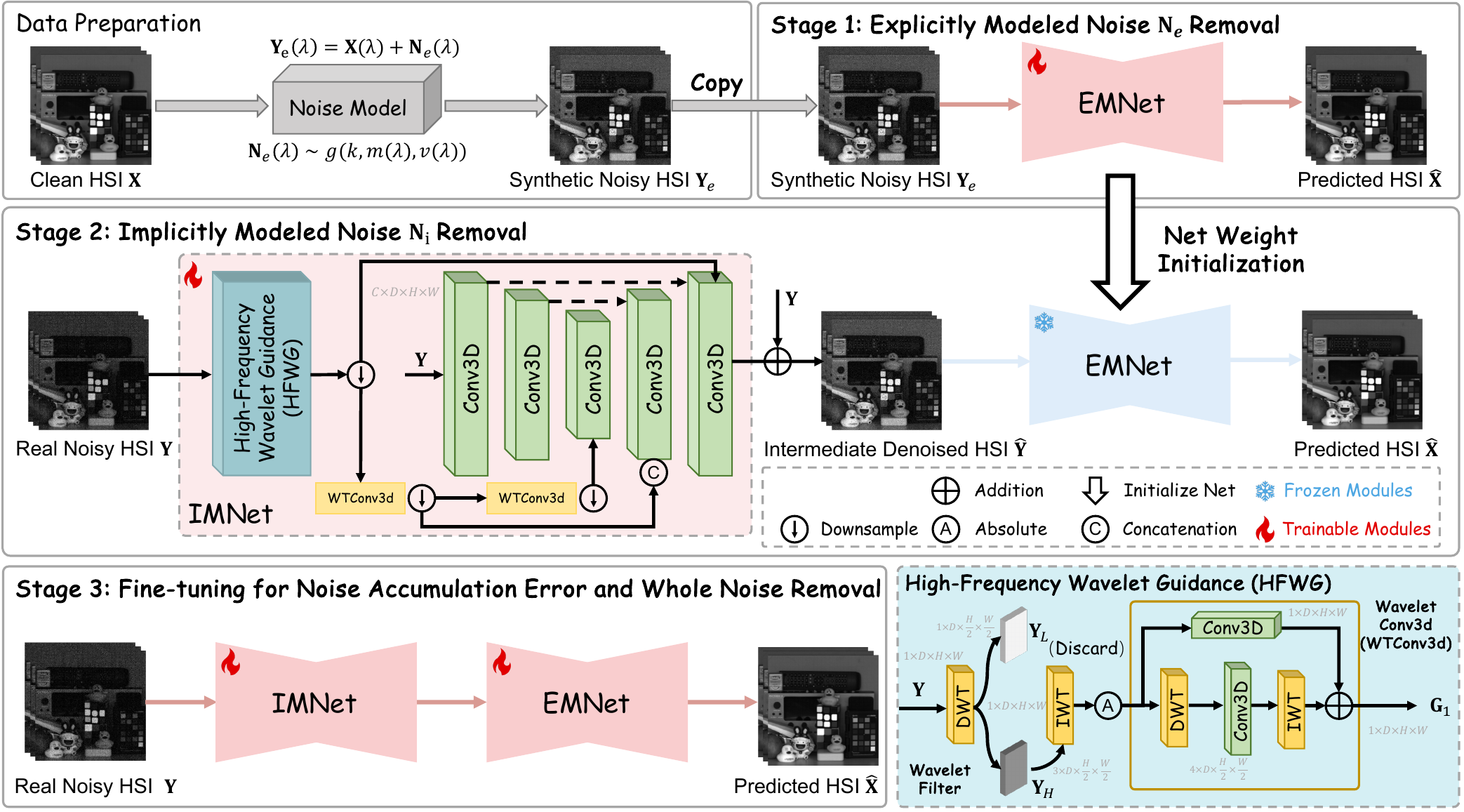}
   \end{center}
	\caption{The overall multi-stage noise-decoupling framework. Stage 1: EMNet removes explicitly modeled noise based on the noise model, guiding the IMNet (Stage 2) to adaptively eliminate implicitly modeled noise. Stage 3: Entire fine-tuning for mitigating noise accumulation errors across stages and effectively removing real-world noise.}
   \label{fig: framework}
   \end{figure*}

\subsection{Noise Modeling}

%

Existing HSI denoising learning-based methods are typically data-driven. Thus, paired clean and noisy data are vital for training and evaluating the performance of denoising networks. There are two main strategies to address this issue. One way is to collect real paired HSI data, for example, 14 pairs of HSIs in~\cite{li2022low} and 59 pairs of HSIs in~\cite{zhang2022guided}. However, there still exists a small number of real paired HSI datasets, and even learning-based methods struggle with the complex data mapping between clean and noisy HSIs.

Another way is to generate more realistic data using noise models, like additive white Gaussian noise~\cite{chang2017hyper}, mixture Gaussian noise~\cite{chen2017denoising}, sensor noise~\cite{acito2011signal,li2025noise}, and impulse noise~\cite{wei20203D}.
Recently, Zhang \etal~\cite{zhang2021hyperspectral,zhang2022guided} propose a noise model that contains more noise sources for the scanning hyperspectral camera and estimate the noise parameters for exactly simulating the original real noise distribution. However, real-world complex noise is challenging to model accurately, leading to discrepancies between synthetic and real data, which adversely affect denoising performance. Thus, we propose a noise-decoupling framework to enhance learnability for real HSI data mappings by separating the noise components into explicitly and implicitly modeled noise.

\section{Multi-Stage Noise-Decoupling Framework}

In this section, we first introduce the formulation and motivation behind our proposed framework. Then, we describe the framework in detail, including both explicitly and implicitly modeled noise removal, along with the multi-stage learning strategy.
The framework is shown in Figure~\ref{fig: framework}. 

\subsection{Formulation and Motivation}

Suppose the clean HSI is denoted as \(\mathbf{X} \in \mathbb{R}^{D\times H \times W}\), where \(H\), \(W\), and \(D\) represent the height, width, and number of spectral bands, respectively. The linear model of the camera between the clean and noisy HSIs can be formulated as:
\begin{equation}
        \mathbf{Y} = \mathbf{X} + \mathbf{N},
    \label{eq: noise}
\end{equation}
where \(\mathbf{Y} \in \mathbb{R}^{D \times H \times W}\) is the noisy HSI. Thus, we aim to remove the noise \(\mathbf{N}\) from the noisy HSI \(\mathbf{Y}\) to restore \(\mathbf{X}\).

As discussed in~\cite{healey1994radiometric, schott2007remote, zhang2022guided}, the noise \(\mathbf{N}\) is commonly categorized into types, such as photon shot noise \(\mathbf{N}_{s}\), readout noise \(\mathbf{N}_{r}\), and stripe noise \(\mathbf{N}_{sp}\).
These components are typically modeled using statistical distributions, such as the Poisson distribution for photon shot noise and the Gaussian distribution for other noise.
However, these idealized models often fail to capture the full complexity of real-world noise. This discrepancy, arising from simplified assumptions or uncharacterized noise sources, is evident both visually and quantitatively. For example, as shown in Figures~\ref{fig: Noise Motivation}(d-f), there are significant regions where the model cannot properly fit the real noise, resulting in a notable PSNR gap between the synthetic and real noisy images.
Thus, the Eq. (\ref{eq: noise}) can be further formulated as:
\begin{equation}
    \mathbf{N} = \mathbf{N}_{e} + \mathbf{N}_{i},
\end{equation}
where \(\mathbf{N}_{e}\) denotes the explicitly modeled noise components, and \(\mathbf{N}_{i}\) denotes the implicitly modeled noise components. 

Therefore, we propose a multi-stage noise-decoupling framework that separates real noise \(\mathbf{N}\) into explicitly modeled noise \(\mathbf{N}_{e}\) and implicitly modeled noise \(\mathbf{N}_{i}\), thereby reducing the difficulty of learning complex mappings. 
Specifically, we leverage an existing noise model to generate synthetic data for training a network, termed EMNet in our framework, to handle the removal of \(\mathbf{N}_{e}\) in the first stage.
Then, to tackle \(\mathbf{N}_{i}\), we propose a high-frequency wavelet guided network, termed IMNet, for the second stage. The design of IMNet is motivated by our observation in Figures~\ref{fig: Noise Motivation}(a-c) that the residual between real and synthetic noise exhibits a high correlation in its high-frequency components.
Leveraging the information-preserving property of the wavelet transform, the wavelet guidance of IMNet adaptively extracts these high-frequency features from the spatial-spectral domain of real noisy HSI.
The primary role of the IMNet is to first remove the complex implicit modeled noise \(\mathbf{N}_{i}\), preparing a cleaner image so that the EMNet can effectively remove the remaining explicit modeled noise \(\mathbf{N}_{e}\).
Finally, the entire network is jointly fine-tuned to mitigate noise accumulation errors across stages, and during the testing phase, the IMNet and EMNet are sequentially applied to effectively remove the real-world noise.

\subsection{Explicitly Modeled Noise Removal}
\label{sec: accurate}

In this stage, we aim to remove noise components, \(\mathbf{N}_{e}\) that can be explicitly modeled. Thus, we generate noisy HSIs based on noise model and pre-train the denoising network to remove noise \(\mathbf{N}_{e}\). Besides, we can leverage this prior knowledge for further complex noise decoupling and removal.

\noindent\textbf{{Data Preparation.}}
Our process begins with the utilization of the physical noise model in~\cite{zhang2022guided} to generate synthetic noisy HSIs. The noisy HSIs based on explicitly modeled noise can be formulated as:
\begin{equation}
   \begin{aligned}
      \mathbf{Y}_{e}(\lambda) &= \mathbf{X}(\lambda) + \mathbf{N}_{e}(\lambda), \\
      \mathbf{N}_{e}(\lambda) &\sim g(k,m(\lambda),v(\lambda)),
   \end{aligned}
   \label{eq: synthetic_noise}
\end{equation}
where \(\mathbf{Y}_{e}(\lambda)\) is the synthetic noisy HSI in \(\lambda\)-th band, and \(\mathbf{N}_{e}(\lambda)\) is the explicitly modeled noise components in \(\lambda\)-th band. \(g(k,m(\lambda),v(\lambda))\) denotes the Poisson/Gaussian distribution with parameters system gain \(k\), mean \(m(\lambda)\), and variance \(v(\lambda)\) for \(\lambda\)-th band. More details about the noise model and calibration are shown in the supplementary material.

\noindent\textbf{{Noise Removal Network.}}
Post generating synthetic noisy HSI, \(\mathbf{Y}_{e}\), we can now pre-train the denoising network to remove explicitly modeled noise \(\mathbf{N}_{e}\). 
We choose recent state-of-the-art networks, which have shown promising results in HSI denoising, as the explicitly modeled noise removal network, termed EMNet in our framework.
Thus, the denoised HSI, \(\hat{\mathbf{X}}\) can be formulated as:
\begin{equation}
   \hat{\mathbf{X}} = f_{EMNet}(\mathbf{Y}_{e};\theta),
\end{equation}
where \(f_{EMNet}(\circ;\theta)\) denotes the EMNet and \(\theta\) is its parameters to be optimized. 

\subsection{Implicitly Modeled Noise Removal}
\label{sec: inaccurate}

In this stage, we aim to remove noise components that cannot be explicitly modeled, \ie, \(\mathbf{N}_{i}\), with the assistance of the prior knowledge from the pre-trained EMNet.
Specifically, we first apply a wavelet filter to extract high-frequency information from the real noisy HSI. Next, we perform wavelet 3D convolution in the wavelet domain, integrating spatial and spectral information to generate multi-scale high-frequency guidance. This multi-scale guidance is then incorporated into the decoder of a 3D U-Net~\cite{ronneberger2015u} as IMNet to guide the noise removal.

Inspired by the high correlation in high-frequency components, we introduce the high-frequency wavelet guidance (HFWG) to adaptively extract high-frequency components to guide the suppression of implicitly modeled noise.
As shown in Figure~\ref{fig: framework}, we first apply a wavelet filter to extract the high-frequency features:
\begin{equation}
   \begin{aligned}
      &\{\mathbf{Y}_{L}, \mathbf{Y}_{H}\} = \text{DWT}(\mathbf{Y}), \\
      &\hat{\mathbf{G}} = \text{abs}(\text{IWT}(\mathbf{Y}_{H})),
   \end{aligned}
\end{equation}
where DWT (Discrete Wavelet Transformation) is used to filter the low- and high-frequency components of the noisy image \(\mathbf{Y}\) and \(\text{IWT}\) denotes the Inverse Wavelet Transform. \(\mathbf{Y}_{L},\mathbf{Y}_{H}\) denotes the low- and high-frequency components. 
Inspired by the design of wavelet 2D convolution in \cite{finder2024wavelet}, we adopt this approach to further adjust high-frequency guidance in the wavelet domain, incorporating 3D spectral-spatial noise information.
The process of wavelet 3D convolution (WTConv3d) can be formulated as:
\begin{equation}
   \begin{aligned}
      &\{\hat{\mathbf{G}}_{L}, \hat{\mathbf{G}}_{H}\} = \text{DWT}(\hat{\mathbf{G}}), \\
      &\mathbf{G}_1 = \text{IWT}(Conv3d(\hat{\mathbf{G}}_{L}, \hat{\mathbf{G}}_{H})\cdot \theta_{w}),
   \end{aligned}
\end{equation}
where \(\theta_{w}\) is the learnable parameters to be optimized. We then utilize the WTConv3d to generate the multi-scale noise feature guidance \(\mathbf{G}_{i}, i\in\{1,2,3\}\) incorporated into the decoder of the backbone, 3D Unet, to suppress the implicitly modeled noise. 
The whole process can be formulated as:
\begin{equation}
   \begin{aligned}
      &\hat{\mathbf{Y}} = f_{IMNet}(\mathbf{Y},\mathbf{G}_{1},\mathbf{G}_{2},\mathbf{G}_{3};\theta),
   \end{aligned}
\end{equation}
where \(f_{IMNet}(\circ;\theta)\) denotes the IMNet to learn a residual for noise removal with the guidance of HFWG.

\begin{figure*}[h]
    \centering
    \hspace{-2mm}
    \begin{minipage}[c]{0.55\textwidth}
        \centering
        \small
        \setlength\tabcolsep{1pt}
        \begin{tabular}{c|c|c|c|c|c}
    \toprule[1.2pt]
    Dataset    & Paired & Sensor  & Bands & Size & Ratios  \\
    \hline
    Urban~\cite{mnih2010learning}& \ding{55} & Hydice  & 210  & 1   & \emph{N/A}       \\ \hline
    LHSI~\cite{li2022low}& \ding{51}  & SPECIM FX10/IQ & 64   & 14   & 15    \\                  
    RealHSI~\cite{zhang2022guided}   & \ding{51} & SOC710-VP & 34  & 59    &   50   \\ \rowcolor{graycolor}
    MEHSI (Ours) & \ding{51}  &  SOC710-VP  & 34  & 303    & 20,50,100      \\\bottomrule[1.2pt]
    \end{tabular}
    \captionof{table}{Summary of existing datasets and our multi-exposure dataset.}
        \label{tab: dataset comparison}
    \end{minipage}%
    \hspace{17mm}
    \begin{minipage}[c]{0.35\textwidth}
        \centering
        \includegraphics[width=0.9\linewidth]{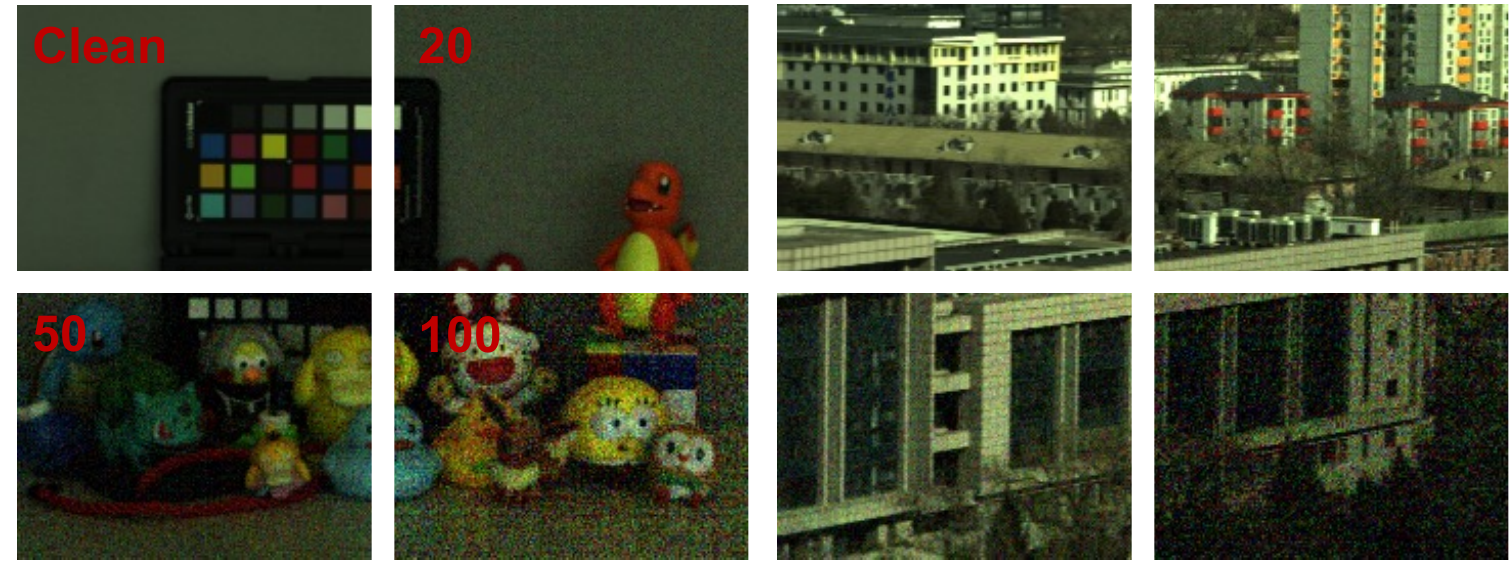}
        \captionof{figure}{Example scenes in MEHSI dataset.}
        \label{fig: dataset}
    \end{minipage}

\end{figure*}

\begin{table*}[htb]
    \centering
    \small
    \setlength{\tabcolsep}{0.12cm}
    \begin{tabular}{c|c|ccc|ccc|ccc|ccc}
        \toprule[1.2pt]
        \multirow{2}{*}{Method} & \multirow{2}{*}{Venue} & \multicolumn{3}{c|}{Ratio 20} & \multicolumn{3}{c|}{Ratio 50} & \multicolumn{3}{c|}{Ratio 100} & \multicolumn{3}{c}{Average} \\
        & & PSNR$\uparrow$ & SSIM$\uparrow$ & SAM$\downarrow$ & PSNR$\uparrow$ & SSIM$\uparrow$ & SAM$\downarrow$ & PSNR$\uparrow$ & SSIM$\uparrow$ & SAM$\downarrow$ & PSNR$\uparrow$ & SSIM$\uparrow$ & SAM$\downarrow$ \\
        \hline
        Noisy & - & 28.18 & 0.919 & 13.451 & 23.59 & 0.777 & 24.677 & 19.51 & 0.543 & 38.028 & 23.76 & 0.746 & 25.385 \\
        QRNN3D & TNNLS'20 & 36.33 & 0.984 & 2.563 & 32.71 & 0.973 & 3.534 & 30.66 & 0.936 & 5.529 & 33.23 & 0.965 & 3.875 \\
        MAC-Net & TGRS'21 & 35.54 & 0.981 & 2.639 & 32.54 & 0.967 & 3.576 & 29.78 & 0.926 & 5.147 & 32.62 & 0.958 & 3.787 \\
        T3SC & NeurIPS'21 & 32.11 & 0.969 & 3.055 & 30.84 & 0.960 & 3.934 & 29.07 & 0.928 & 4.595 & 30.67 & 0.953 & 3.861 \\
        GRNet & TGRS'21 & 35.08 & 0.970 & 2.953 & 32.55 & 0.944 & 3.821 & 29.02 & 0.878 & 4.821 & 32.22 & 0.931 & 3.865 \\
        SST & AAAI'23 & 37.23 & 0.986 & 2.430 & 34.98 & 0.979 & 3.367 & 31.68 & 0.952 & 5.147 & 34.63 & 0.973 & 3.648 \\
        SERT & CVPR'23 & \underline{37.59} & \underline{0.986} & 2.361 & \underline{35.27} & \underline{0.980} & 3.351 & 31.68 & 0.955 & 4.989 & 34.85 & 0.974 & 3.567 \\
        HSDT & ICCV'23 & 35.99 & 0.984 & 2.654 & 32.78 & 0.973 & 3.672 & 31.37 & 0.946 & 5.613 & 33.38 & 0.967 & 3.980 \\
        TDSAT & TGRS'24 & 37.45 & \underline{0.986} & 2.636 & 34.72 & 0.978 & 3.229 & \underline{32.56} & \underline{0.959} & 4.643 & \underline{34.91} & \underline{0.975} & 3.503 \\
        HIRDiff & CVPR'24 & 31.80 & 0.959 & 7.412 & 29.59 & 0.928 & 8.922 & 25.90 & 0.805 & 13.506 & 29.09 & 0.897 & 9.946 \\
        VolFormer & CVPR'25 & 37.37 & \underline{0.986} & \underline{2.198} & 34.90 & \underline{0.980} & \underline{2.867} & 32.40 & 0.955 & \underline{4.095} & 34.89 & 0.974 & \underline{3.053} \\ \rowcolor{graycolor}
        Ours (TDSAT) & - & \textbf{38.44} & \textbf{0.988} & \textbf{1.822} & \textbf{36.64} & \textbf{0.983} & \textbf{2.149} & \textbf{34.01} & \textbf{0.972} & \textbf{2.714} & \textbf{36.36} & \textbf{0.981} & \textbf{2.228} \\
        \bottomrule[1.2pt]
    \end{tabular}
    \caption{Averaged results of different methods under different noise levels on the MEHSI dataset. The best and second-best results are highlighted in bold and underlined, respectively.}
    \label{tab: result on ME Real HSI dataset}
\end{table*}

\subsection{Multi-Stage Learning Strategy}

\noindent\textbf{{Pre-training for Explicitly Modeled Noise Removal.}}
We apply the Charbonnier loss~\cite{charbonnier1994two} to constrain the EMNet to focus on explicitly modeled noise removal:
\begin{equation}
   \mathcal{L}_{c} = \sqrt{||\mathbf{X} - \hat{\mathbf{X}}||^{2} + \epsilon^{2}},
\end{equation}
where \(\mathbf{X}\) is the ground truth clean image, \(\hat{\mathbf{X}}\) is the predicted clean image, and \(\epsilon=10^{-3}\) is a constant.

\noindent\textbf{{Learning for Implicitly Modeled Noise Removal.}}
Due to the unknown noise distribution, we cannot acquire the ground truth only with explicitly modeled noise, but we can utilize the pre-trained EMNet as a discriminator with prior knowledge to distinguish the distribution of explicitly modeled noise based on real paired data. Thus, we freeze the EMNet during this training stage and also apply \(\mathcal{L}_{c}\) loss function to constrain the learning process to update the parameters in IMNet. Besides, to further constrain the output of IMNet, we adopt the Kullback-Leibler divergence loss function to enforce distribution consistency between its output and the corresponding synthetic data:
\begin{equation}
   \mathcal{L}_{k} = \sum_{i} p(\mathbf{Y}_i) \log \frac{p(\mathbf{Y}_i)}{q(\hat{\mathbf{Y}}_i)},
\end{equation}
where \(p(\mathbf{Y}_i)\) is the actual noise distribution and \(q(\hat{\mathbf{Y}}_i)\) is the predicted noise distribution. The \(p(\mathbf{Y}_i)\) can be generated by the explicitly modeled noise during the training process, \ie, online, or before the training process, \ie, offline.

\noindent\textbf{{Fine-tuning for Noise Accumulation Error and Whole Noise Removal.}}
To further remove whole noise and eliminate the accumulated noise errors from each previous stage, we perform joint fine-tuning on the pre-trained IMNet and EMNet.
Except for the above loss functions, we also apply the spectral consistency loss to guide the network to focus on spectral fidelity:
\begin{equation}
   \mathcal{L}_{s} = 1- \frac{1}{N} \sum_{i} \frac{\mathbf{X}_i \cdot \hat{\mathbf{X}}_i}{\|\mathbf{X}_i\|^{2} \| \hat{\mathbf{X}}_i \|^{2}},
\end{equation}
where \( N \) is the number of pixels in the image. 
The overall loss function is defined as follows:
\begin{equation}
   \mathcal{L} = \mathcal{L}_{c} + \lambda_{k} \mathcal{L}_{k} + \lambda_{s} \mathcal{L}_{s},
\end{equation}
where \( \lambda_{k} \) and \( \lambda_{s} \) are hyperparameters.

\section{Experiments}

In this section, we first introduce the datasets, implementation details, metrics for quantitative evaluation, and compared methods. Then, we provide quantitative and qualitative results. We further conduct ablation studies and discussions to evaluate the effectiveness of the proposed approach.

\subsection{Datasets and Experimental Settings}

\noindent{\textbf{Datasets.}}
Real experiments are conducted on two datasets: the RealHSI dataset~\cite{zhang2022guided} and our collected Multi-Exposure real HSI denoising (MEHSI) dataset.
The \textbf{RealHSI} dataset consists of 59 paired clean and noisy images, each captured under a single exposure ratio. Each HSI contains 34 bands from 400 $nm$ to 700 $nm$ with a size of $696\times520$ pixels. 
Due to the limited data volumes and limited noise level in the RealHSI dataset, we additionally capture a larger real dataset, \textbf{MEHSI}, with various noise levels.
We capture 101 indoor and outdoor scenes with 3 different noise levels, totaling 303 pairs by the SOC710-VP hyperspectral camera.
The image pre-processing is the same as that of the RealHSI dataset to obtain paired HSIs with 34 bands. The comparisons of datasets are shown in Table~\ref{tab: dataset comparison} and example scenes of MEHSI dataset are shown in Figure~\ref{fig: dataset}. More details can be found in the supplementary material.

\begin{table}[t]
    \begin{center}
    \small
    
    \setlength{\tabcolsep}{0.25cm}
    \begin{tabular}{c|c|ccc}
    \toprule[1.2pt]
    Method & Venue & PSNR$\uparrow$ & SSIM$\uparrow$ & SAM$\downarrow$ \\ \hline
    Noisy & - & 23.26 & 0.760 & 17.329 \\
    QRNN3D & TNNLS'20 & 30.42 & 0.953 & 3.939 \\
    MAC-Net & TGRS'21 & 28.82 & 0.936 & 5.227 \\
    T3SC & NeurIPS'21 & 27.79 & 0.901 & 4.220 \\
    GRNet & TGRS'21 & 28.44 & 0.909 & 3.956 \\
    SST & AAAI'23 & 28.50 & 0.948 & 3.885 \\
    SERT & CVPR'23 & 29.01 & 0.939 & \underline{3.202} \\
    HSDT & ICCV'23 & \underline{31.24} & \underline{0.958} & 3.751 \\
    TDSAT & TGRS'24 & 30.70 & 0.958 & 3.241 \\
    HIRDiff & CVPR'24 & 30.34 & 0.943 & 4.923 \\
    VolFormer & CVPR'25 & 29.33 & 0.930 & 3.231 \\ \rowcolor{graycolor}
    Ours (HSDT) & - & \textbf{32.31} & \textbf{0.967} & \textbf{2.742} \\
    \bottomrule[1.2pt]
    \end{tabular}
    \caption{Averaged results of different methods on the RealHSI dataset. The best and second-best results are highlighted in bold and underlined, respectively. }
    \label{tab: result on RealHSI dataset}
    \end{center}
\end{table}

\begin{table*}[th]
   \setlength{\tabcolsep}{0.21cm}
      \centering
      \small
      \begin{tabular}{c|c|ccc|ccc|ccc}
      \toprule[1.2pt]
      \textbf{Strategy} & \textbf{Loss} & Pre-trained & Online & Offline  & EMNet & IMNet & HFWG & PSNR$\uparrow$ & SSIM$\uparrow$ & SAM$\downarrow$ \\ \hline
      \multirow{2}{*}{End-to-End}
      & $\mathcal{L}_{c}$ & \ding{55} & \ding{55} & \ding{55} & \ding{51} & \ding{55} & \ding{55} & 34.91 & 0.975 & 3.503 \\
      & $\mathcal{L}_{c}+\mathcal{L}_{s}$ & \ding{55} & \ding{55} & \ding{55} &  \ding{51} & \ding{55} & \ding{55} & 35.33 & 0.975 & 2.471 \\ 
      & $\mathcal{L}_{c}+\mathcal{L}_{s}$ & \ding{55} & \ding{55} & \ding{55} &  \ding{51} & \ding{51} & \ding{55} & 35.56 & 0.976 & 2.459 \\ \hline 
      \multirow{5}{*}{Multi-Stage}
      & $\mathcal{L}_{c}$ & \ding{51} & \ding{55} & \ding{55} & \ding{51} & \ding{51} & \ding{55} & 35.66 & \underline{0.978} & 3.240 \\
      & $\mathcal{L}_{c}+\mathcal{L}_{s}$ & \ding{51} & \ding{55} & \ding{55} & \ding{51} & \ding{51} & \ding{55} & 35.83 & 0.977 & \textbf{2.228} \\
      & $\mathcal{L}_{c}+\mathcal{L}_{k}+\mathcal{L}_{s}$ & \ding{51} & \ding{51} & \ding{55} & \ding{51} & \ding{51} & \ding{55} & \underline{36.16} & 0.977 & \underline{2.253} \\
      & $\mathcal{L}_{c}+\mathcal{L}_{k}+\mathcal{L}_{s}$ & \ding{51} & \ding{55} & \ding{51} & \ding{51} & \ding{51} & \ding{51} (3K Params) & 36.04 & 0.976 & 2.364 \\
      \rowcolor{graycolor}
      (Ours) & $\mathcal{L}_{c}+\mathcal{L}_{k}+\mathcal{L}_{s}$ & \ding{51} & \ding{51} & \ding{55} & \ding{51} & \ding{51} & \ding{51} (3K Params) & \textbf{36.36} & \textbf{0.981} & \textbf{2.228} \\
      \bottomrule[1.2pt]
      \end{tabular}
      \caption{Ablation studies on the learning strategy and network designs towards higher performance.}
      \label{tab: ablation strategy}
\end{table*}

\noindent{\textbf{Implementation Details.}}
We implement the proposed framework with Pytorch~\cite{paszke2019pytorch}. In detail, we crop overlapped
$128\times128$ spatial regions from the paired data and augment them by random flipping and/or rotation.
Following the settings in SERT~\cite{li2023spectral}, we randomly choose 44 HSIs for training and the remaining 15 for testing in the RealHSI dataset. For MEHSI dataset, we randomly select 273 pairs for training and the remaining 30 for testing.
The models are trained with Adam~\cite{kingma2014adam} ($\beta_{1}$ = 0.9 and $\beta_{2}$ = 0.999) for 200 epochs in RealHSI and 400 epochs in our MEHSI dataset. The initial learning rate and batch size are set to $1\times10^{-4}$ and 1, respectively. 
Experiments are conducted on a single NVIDIA RTX 4090 GPU.

\noindent{\textbf{Evaluation Metrics.}}
We utilize three quantitative image quality metrics: Peak Signal-to-Noise Ratio (PSNR), Structural Similarity Index (SSIM), and Spectral Angle Mapping (SAM). 
Higher PSNR and SSIM values, combined with lower SAM values, indicate better performance.

\noindent{\textbf{Compared Methods.}}
We compare our proposed framework against SOTA deep learning-based HSI denoising methods, including QRNN3D~\cite{wei20203D}, MAC-Net~\cite{xiong2021mac}, T3SC~\cite{bodrito2021trainable}, GRNet~\cite{cao2021deep}, SST~\cite{li2023spatial}, SERT~\cite{li2023spectral}, HSDT~\cite{lai2023hybrid}, TDSAT~\cite{zhang2024three}, HIRDiff~\cite{pang2024hir}, and VolFormer~\cite{yu2025volformer}. 
All compared methods are trained with the same settings as ours for fair comparison.

\subsection{Results on MEHSI Dataset}

\noindent\textbf{Quantitative Comparison.}
We show the quantitative results of various methods on the MEHSI dataset in Table~\ref{tab: result on ME Real HSI dataset}. Among all methods, HIRDiff is an unsupervised method that can partially denoise compared to the original noisy image. However, its performance is suboptimal when denoising real noise, especially under varying noise levels. In contrast, our proposed method, based on TDSAT, achieves an average improvement of at least 1.45 dB in PSNR and 0.825 in SAM across multiple noise levels, compared to other learning-based approaches. Notably, our method effectively removes noise across different noise levels, demonstrating stability in real-world complex noise scenarios. 
This highlights the effectiveness of our framework.

\noindent\textbf{Qualitative Comparison.}
We visualize the results in the upper part of Figure~\ref{fig: qualitative results}, where the exposure ratio of the input noisy image is 50. Most methods still leave significant residual noise and blur, particularly in the results of T3SC and GRNet. HIRDiff exhibits over-smoothing, which damages texture details. In comparison, our method effectively removes noise, resulting in a smaller error map. The superior performance is primarily attributed to our multi-stage noise-decoupling approach, which effectively reduces noise complexity, with the multi-stage learning strategy to enhance the learning ability of networks.

\subsection{Results on RealHSI Dataset}

\noindent\textbf{Quantitative Comparison.}
We conduct experiments on the RealHSI dataset to evaluate the performance of our proposed method, as shown in Table~\ref{tab: result on RealHSI dataset}. 
It can be observed that, compared to the second-best method, HSDT, our method achieves higher PSNR and SSIM and lower SAM values, demonstrating superior denoising performance.

\begin{figure*}[t]
   \begin{center}
   \includegraphics[width=1\linewidth]{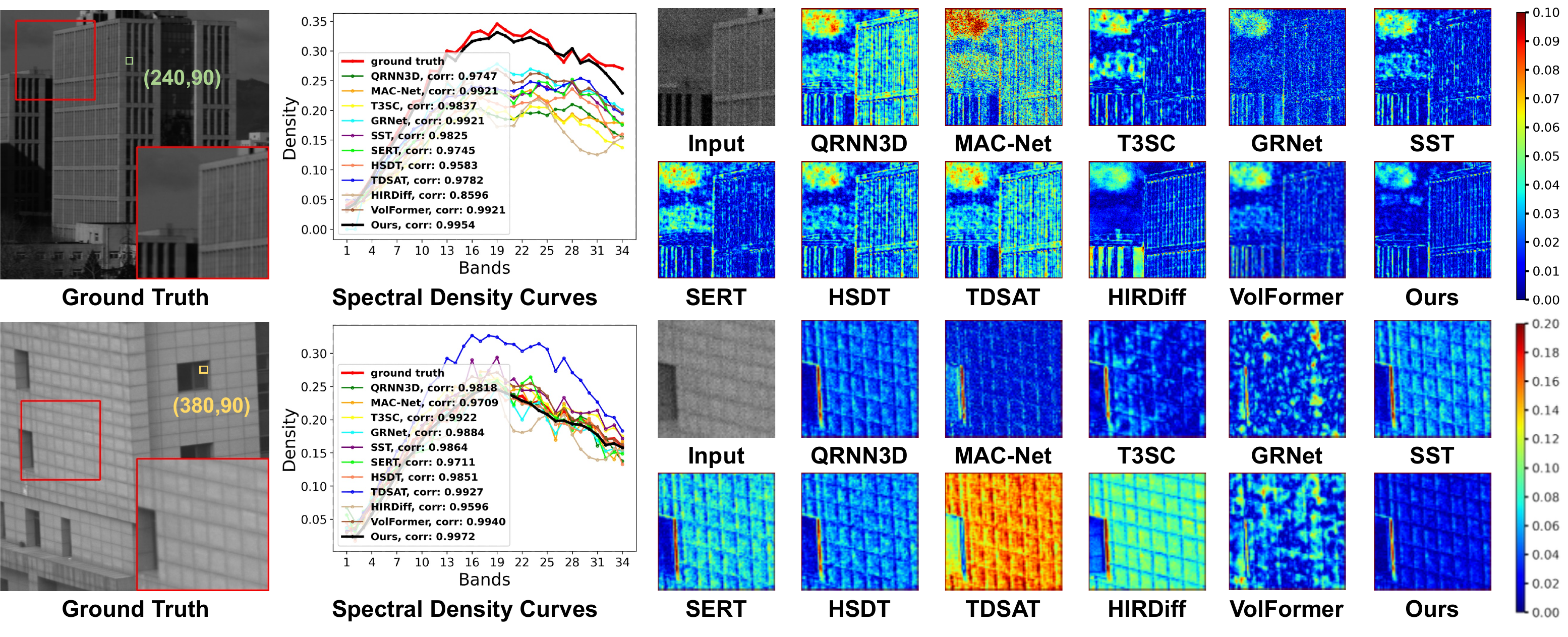}
   \end{center}
	\caption{Visual quality comparisons of sample scene on MEHSI (up) and RealHSI (bottom) datasets with the spectral band in 510 $nm$, and 550 $nm$, respectively. The `corr' of spectral density curves in the color box represents the correlation coefficient.
	}
   \label{fig: qualitative results}
   \end{figure*}

\noindent\textbf{Qualitative Comparison.}
A visual comparison of results for a scene is presented in the lower part of Figure~\ref{fig: qualitative results}. As shown, the comparison methods either retain more residual noise or overly smooth the image, leading to a larger error map. In contrast, our method achieves more effective noise removal while preserving texture details, making the result closer to the reference image. 

\begin{table}[t]
   \setlength{\tabcolsep}{0.09cm}
     \centering
     \small
   
      \begin{tabular}{c|cc|cc|cc}
      \toprule[1.2pt]
      \textbf{Model} & P(M) & F(G) & PSNR$\uparrow$ & $\Delta$$\uparrow$ & SAM$\downarrow$ & $\Delta$$\downarrow$ \\ \hline
      HSDT & 0.52 & 207 & 33.38 & - & 3.980 & - \\
      HSDT(\textit{Pt}) & 0.52 & 207 & 33.46 & \textgreen{+0.08} & 3.989 & \textred{+0.009} \\
      HSDT-L & 0.98 & 389 & 33.60 & \textgreen{+0.22} & 3.875 & \textgreen{-0.105} \\ \rowcolor{graycolor}
      Ours(HSDT) & 0.62 & 260 & \textbf{34.18} & \textgreenbf{+0.80} & \textbf{2.443} & \textgreenbf{-1.537} \\ \hline
      VolFormer & 2.41 & 109 & 34.89 & - & 3.053 & - \\
      VolFormer(\textit{Pt}) & 2.41 & 109 & 35.33 & \textgreen{+0.44} & 2.820 & \textgreen{-0.090} \\
      VolFormer-L & 3.24 & 263 & 35.00 & \textgreen{+0.11} & 3.089 & \textred{+0.036} \\ \rowcolor{graycolor}
      Ours(VolFormer) & 2.80 & 320 & \textbf{36.06} & \textgreenbf{+1.17} & \textbf{2.164} & \textgreenbf{-0.889} \\ \hline
      TDSAT & 1.09 & 501 & 34.91 & - & 3.503 & - \\
      TDSAT(\textit{Pt}) & 1.09 & 501 & 35.26 & \textgreen{+0.35} & 3.251 & \textgreen{-0.252}  \\
      TDSAT-L & 1.69 & 774 & 35.16 & \textgreen{+0.25} & 3.399 & \textgreen{-0.104} \\
      TDSAT(HSDT) & 1.62 & 708 & 35.60 & \textgreen{+0.69} & 2.343 & \textgreen{-1.160} \\ \rowcolor{graycolor}
      Ours(TDSAT) & 1.31 & 621 & \textbf{36.36} & \textgreenbf{+1.45} & \textbf{2.228} & \textgreenbf{-1.275} \\
      \bottomrule[1.2pt]
      \end{tabular}
      \caption{Comparison of the model complexity. `P(M)' means Parameters(M) and `F(G)' means Flops(G). `\textit{Pt}' denotes that the model is pre-trained on synthetic data, and `$*$-L' means a larger model with more parameters.}
      \label{tab: ablation params}
\end{table}

\subsection{Ablation Study}
\label{sec: ablation}

\noindent\textbf{{Learning Strategy.}} 
As shown in Table~\ref{tab: ablation strategy}, we conduct ablation studies on the multi-stage learning strategy based on the network, TDSAT.
First, we adopt an end-to-end learning strategy, applying \(\mathcal{L}_c\) and \(\mathcal{L}_s\) loss functions as constraints. We observe that \(\mathcal{L}_s\) can effectively improve the SAM metric, enhancing spectral fidelity with a slight increase in PSNR performance. Next, we pre-train TDSAT and incorporate it into our framework as EMNet, resulting in at least a 0.3 dB improvement in PSNR. 
Additionally, as shown in Table~\ref{tab: ablation params}, directly using the `\textit{Pt}', \ie, pre-training strategy, leads to performance gains, and incorporating it into our framework can make consistent improvement.
Subsequently, we employ \(\mathcal{L}_k\) to constrain the training of IMNet. 
By comparing noise synthetic strategies, we use the online strategy to enhance the learning performance of the framework.
More details are shown in the supplementary material.

\begin{figure}[t]
    \centering
    \begin{minipage}[t]{0.22\textwidth}
        \centering
        \small
        
        \setlength\tabcolsep{3pt}
    \begin{tabular}{l|cc}
    \toprule[1.2pt]
    Methods   & PSNR$\uparrow$ & SSIM$\uparrow$  \\ \hline
	HSDT & 34.62 & 0.940   \\
	TDSAT & 36.38 & 0.942   \\
	VolFormer & 34.42 & 0.925  \\ \rowcolor{graycolor}
	Ours &  \textbf{37.09} & \textbf{0.946}   \\ \bottomrule[1.2pt]
        \end{tabular}
        \captionof{table}{Comparison results on the LHSI dataset.} 
        \label{tab: comparison on LHSI}
    \end{minipage}
    \hspace{3mm}
    \begin{minipage}[t]{0.22\textwidth}
        \centering
        \small
        
        \setlength\tabcolsep{3pt}
    \begin{tabular}{l|cc}
    \toprule[1.2pt]
    \textbf{($\lambda_k$, $\lambda_s$)}   & PSNR$\uparrow$ & SSIM$\uparrow$  \\ \hline
	(0.01,100) & 31.41 & 0.965 \\ \rowcolor{graycolor}
        	(0.01,10) & \textbf{32.31} & \textbf{0.967} \\
        (0.01,1) & 31.92 & 0.967  \\
        (0.1,10) & 31.06 & 0.965 \\ \bottomrule[1.2pt]
        \end{tabular}
        \captionof{table}{Analysis on the effect of hyperparameters.} 
        \label{tab: hyperparameters}
    \end{minipage}
\end{figure}

\noindent\textbf{{Network Design.}} 
As shown in Table~\ref{tab: ablation strategy}, we first conduct an ablation study on the modules of our framework. The results demonstrate that incorporating the IMNet in a multi-stage configuration enhances denoising performance, even within an end-to-end training. Subsequently, we perform a further ablation on the High-Frequency Wavelet Guidance (HFWG). The results show that with a slight increase in parameters, PSNR improves by 0.2 dB. 
Additionally, in Table~\ref{tab: ablation params}, using HSDT as the IMNet outperforms the single TDSAT, which validates our multi-stage noise-decoupling and learning strategy. Furthermore, our proposed IMNet with HFWG surpasses the larger HSDT model, demonstrating the effectiveness of our network design for noise removal.

\subsection{Discussion}
\label{sec: discussion}

\noindent\textbf{Spectral Consistency.}
In Figure~\ref{fig: qualitative results}, we plot the spectral density curves corresponding to the small green-boxed and yellow-boxed region on MEHSI and RealHSI datasets. The high correlation and significant overlap between our curve and the ground truth demonstrate the effectiveness of our method in preserving spectral consistency.

\noindent\textbf{Model Complexity.}
As shown in Table~\ref{tab: ablation params}, the `model-L' refers to a variant with a larger parameter count. 
The results show that while a larger model achieves better performance compared to its smaller counterpart, our proposed framework based on these smaller models, with fewer parameters and computation, still outperforms the larger model, highlighting the effectiveness of our framework. 

\noindent\textbf{Generalization.}
Beyond demonstrating the generalization of the framework with various backbones on the MEHSI dataset (Table~\ref{tab: ablation params}), we conduct further experiments on the LHSI dataset with other sensors, spectral resolution, and acquisition conditions (Table~\ref{tab: comparison on LHSI}). The results also outperform competing methods, which underscores the generalization and generality of our model. It demonstrates that our framework is generic, exhibiting adaptability with diverse HSI hardware setups, and is not restricted to a specific sensor.
More details about the generalization discussion on noise models are shown in the supplementary material.

\noindent\textbf{Hyperparameters.}
As indicated in Table~\ref{tab: hyperparameters}, performance exhibits sensitivity to the $\lambda_k$ and $\lambda_s$, leading to variations. Based on the experimental results, we choose $\lambda_k=0.01$ and $\lambda_s=10$ in our experiments.

\section{Conclusion}
In this paper, we propose a multi-stage noise-decoupling framework for real hyperspectral image denoising. The framework integrates prior knowledge of a physical noise model to effectively decouple complex noise into explicitly and implicitly modeled components, thereby reducing the learning difficulty at each stage.
We introduce high-frequency wavelet guidance for adaptive high-frequency feature extraction, enabling our network to focus on eliminating implicitly modeled noise by leveraging the explicit noise patterns learned in the previous stage.
We develop a multi-stage learning strategy that comprises separate pre-training and joint fine-tuning of the networks to ensure effective training and mitigate error accumulation
The quantitative and qualitative results on real datasets demonstrate that our method achieves superior denoising performance compared with state-of-the-art methods.

\section{Acknowledgments}
This work was supported by the National Natural Science Foundation of China (62331006 and 62171038), and the Fundamental Research Funds for the Central Universities.

\bibliography{aaai2026}

\end{document}